\documentclass[letterpaper]{article} 
\usepackage[submission]{aaai25}  
\usepackage{times}  
\usepackage{helvet}  
\usepackage{courier}  
\usepackage[hyphens]{url}  
\usepackage{graphicx} 
\usepackage{natbib}  
\usepackage{caption} 
\usepackage{algorithm}
\usepackage{algpseudocode} 
\usepackage{graphicx}
\usepackage{booktabs}
\usepackage{multirow} 
\usepackage{tabularx} 
\usepackage{adjustbox}
\usepackage{color, colortbl}
\definecolor{gray}{gray}{0.1}
\usepackage{nicematrix,tikz}
\usepackage{lipsum}
\usepackage{amssymb}
\usepackage{subcaption}

\newcolumntype{Y}{>{\centering\arraybackslash}X}

\usepackage{newfloat}
\usepackage{listings}
\DeclareCaptionStyle{ruled}{labelfont=normalfont,labelsep=colon,strut=off} 
\floatstyle{ruled}
\newfloat{listing}{tb}{lst}{}
\floatname{listing}{Listing}
\title{SSTD: Stripe-Like Space Target Detection Using Single-Point Weak Supervision}
\author{%
Zijian Zhu,\textsuperscript{\rm 1,2,3}
Ali Zia,\textsuperscript{\rm 4}
Xuesong Li,\textsuperscript{\rm 4}
Bingbing Dan,\textsuperscript{\rm 1,2}
Yuebo Ma,\textsuperscript{\rm 1,2,3}
Enhai Liu,\textsuperscript{\rm 1,2,3}
Rujin Zhao\textsuperscript{\rm 1,2,3}%
}
\affiliations{%
\textsuperscript{\rm 1}Institute of Optics and Electronics, Chinese Academy of Sciences, Chengdu, China\\
\textsuperscript{\rm 2}Key Laboratory of Science and Technology on Space Optoelectronic Precision Measurement, Chinese Academy of Sciences, Chengdu, China\\
\textsuperscript{\rm 3}University of Chinese Academy of Sciences, Beijing, China\\
\textsuperscript{\rm 4}College of Science, Australian National University, Australia
}

\usepackage{bibentry}

\begin{document}

\maketitle


\begin{abstract}
Stripe-like space target detection (SSTD) plays a key role in enhancing space situational awareness and assessing spacecraft behaviour. This domain faces three challenges: the lack of publicly available datasets, interference from stray light and stars, and the variability of stripe-like targets, which makes manual labeling both inaccurate and labor-intensive. In response, we introduces `AstroStripeSet', a pioneering dataset designed for SSTD, aiming to bridge the gap in academic resources and advance research in SSTD. Furthermore, we propose a novel teacher-student label evolution framework with single-point weak supervision, providing a new solution to the challenges of manual labeling. This framework starts with generating initial pseudo-labels using the zero-shot capabilities of the Segment Anything Model (SAM) in a single-point setting. After that, the fine-tuned StripeSAM serves as the teacher and the newly developed StripeNet as the student, consistently improving segmentation performance through label evolution, which iteratively refines these labels. We also introduce `GeoDice', a new loss function customized for the linear characteristics of stripe-like targets. Extensive experiments show  that our method matches fully supervised approaches, exhibits strong zero-shot generalization for diverse space-based and ground-based real-world images, and sets a new state-of-the-art (SOTA) benchmark. Our AstroStripeSet dataset and code will be made publicly available.
\end{abstract}

\section{Introduction}
\label{sec:intro}

\begin{figure}[ht]
  \centering 
  \includegraphics[width=1.0\linewidth]{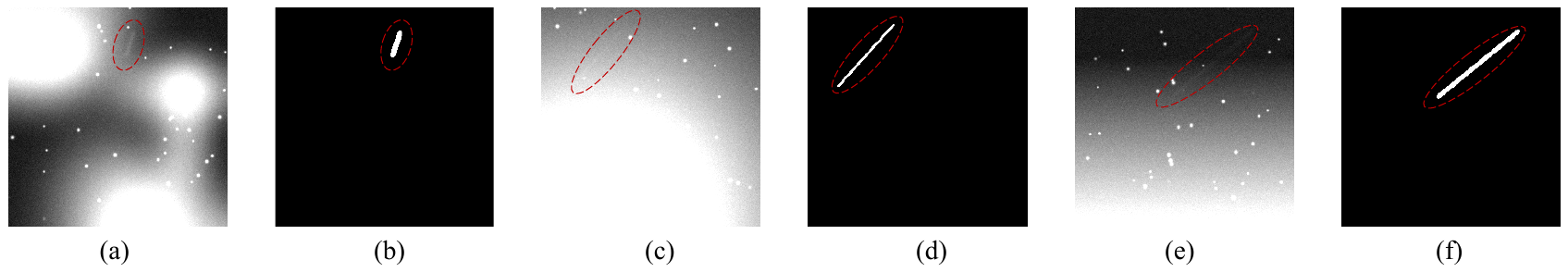}
  \caption{Challenges in the SSTD task. (a), (c) and (e) show space images with different stray light, while (b), (d) and (f) are the stripe-like target's ground truths. } 
  \label{fig:fig1} 
\end{figure}

In the evolving landscape of global space activities, the exponential increase in space targets, encompassing space debris, spacecraft, and asteroids of significant astronomical interest, underscores the critical importance of space target detection. 
It provides key technical support for debris mitigation, long-distance laser communications, and asteroid exploration \cite{liu2020topological,yang2020hcnn}. High-precision optical telescopes, necessitated by the vast distances between space targets and observational equipment, have become the preferred instruments \cite{lin2021new}. These telescopes capture space targets as either stripe-like or point-like, depending on the operational mode (target tracking or star tracking) and exposure duration \cite{liu2020space}. Extended exposure times in star tracking mode enhance space target detection, presenting them as stripe-like due to their motion \cite{lin2021new}. However, this advantage is counterbalanced by an increased vulnerability to space stray light during long exposures, which significantly compromises the signal-to-noise ratio (SNR) of these stripe-like targets, as shown in Figure \ref{fig:fig1}. 
This domain faces several challenges:



\textit{Dataset Availability}: 
The lack of a large-scale, open source dataset has been a significant obstacle in advancing SSTD research. The difficulty and high cost of acquiring numerous real space images have led to a shortage of dataset for thoroughly benchmarking detection algorithms.

\textit{Space Noise}: The SSTD task is significantly challenged by the low SNR, due to interference from space stray light and numerous stars \cite{chen2023star}, as shown in Figure \ref{fig:fig1}. This interference complicates the labeling process for data-driven methods~\cite{liu2023multi}.

\textit{Target Variability}: The variability of stripe-like targets, including differences in position, scale, direction, and brightness distribution, poses a challenge to traditional detection methods \cite{lin2021new,diprima2018efficient}, necessitating more adaptable and robust approaches.

To address the scarcity of public datasets for the SSTD task, we developed the AstroStripeSet dataset. To the best of our knowledge, this dataset is the first of its kind, featuring synthesized images under various stray light scenarios with precise labels in multiple formats. This initiative fills a critical gap in SSTD by providing a foundational resource for benchmarking and advancing research in this field. 

Notably, foundation models such as DINO \cite{oquab2023dinov2} and SAM \cite{kirillov2023segment} have demonstrated remarkable zero-shot capabilities. These models are pre-trained on extensive datasets, enabling them to learn general pattern features and exhibit strong performance in downstream tasks without requiring task-specific training. To address the poor generalization of the existing methods for SSTD, we explore the potential of foundation models. The emergence of these models, particularly those with enhanced scene generalization capabilities like SAM, offers a promising avenue for the SSTD task. To address the issues of inaccurate and time-consuming manual labeling, we focus on the single-point prompt segmentation capability of SAM. However, adapting these models to specialized domains such as space images remains a challenge, necessitating fine-tuning strategies to effectively leverage their capabilities \cite{huang2023segment,ren2024segment,yao2023sam}.
Recent advances in single-point weak supervision \cite{ying2023mapping, Li_2023_ICCV} have highlighted its potential to significantly reduce labeling efforts, but their effectiveness has been demonstrated only on small targets. 

Building on these insights, this work pioneers the application of a single-point weakly supervised strategy for the SSTD task. Specifically, we propose the teacher-student framework for label evolution, which first utilizes the zero-shot capabilities of SAM to generate initial pseudo-labels from single-point labels. The teacher model StripeSAM is fine-tuned through low-rank adaptation with the filtered pseudo-labels and can segment more faint texture stripes. The refined pseudo-labels are then employed to train the student model, StripeNet, to distill knowledge from the teacher model and enlarge the search space for better pseudo-labels. The introduction of the customized StripeNet provides new insights into stripes segmentation for the StripeSAM. This iterative teacher-student mechanism facilitates label evolution and improves SSTD accuracy. Additionally, we proposed a new loss function called GeoDice, tailored to the straight line shape characteristics of stripe-like targets, which greatly improved the teacher-student model's ability to locate stripe-like patterns and suppress irrelevant features during the label evolution process.

Our contributions are manifold, outlined as follows:

\begin{itemize}
\item 
We introduce the pioneering AstroStripeSet dataset, which establishes the first benchmark for SSTD research.

\item We develop an innovative teacher-student label evolution framework for single-point weak supervision, which gets rid of reliance on pixel-level labels and improves detection accuracy for variable, low SNR stripe-like targets.

\item We propose GeoDice, a novel loss function tailored for stripe-like targets, to help the framework more accurately locate stripe-like targets and suppress false alarms.

\item Extensive experiments show that our framework achieves SOTA performance, matches fully supervised methods, and exhibits strong zero-shot generalization on various space-based and ground-based real-world images.
\end{itemize}

\section{Related Work}
\label{sec:bc}

This section reviews detection techniques in complex backgrounds, categorizing them into geometric feature-based, pattern-based, and data-driven approaches.

\subsection{Geometric Feature-Based Approaches}
Geometric feature-based methods often use parameter space accumulation or integral transformations to detect stripe-like targets in images.  
\cite{hickson2018fast} combined matched filtering with an improved Radon transform for SSTD. \cite{jiang2022space, jiang2022automatic} used enhanced median filtering and Hough transform for extracting stripes against non-uniform backgrounds. \cite{liu2020space1} and \cite{xi2016space} applied time-index filtering methods to remove stars and noise, followed by multi-frame state transfer and multi-stage hypothesis testing for SSTD. Despite their succes, these methods struggle with background noise and variable stripe-like targets.

\subsection{Pattern-Based Approaches}
Pattern-based approaches, akin to geometric feature-based methods, commence with filtering algorithms to mitigate space noise and stars. The distinction lies in their extraction techniques, where multiple stripe-matching templates of various orientations are used to detect stripe-like targets. \cite{virtanen2016streak} introduced a multi-window geometric feature analysis for automatic SSTD, requiring motion prior information.  \cite{sara2017faint} explored stripe pattern matching filters and faint stripes detection methods based on stripe pattern distribution, respectively. A recent work by \cite{lin2021new} attempted to detect stripe-like targets using  pattern clustering. However, it cannot detect faintly textured stripe-like targets under intense stray light noise.

\subsection{Data-Driven Approaches}
Integrating convolutional neural networks (CNNs) into SSTD marks a shift towards data-driven methods, automating the learning of stripe features from extensively labeled images. \cite{jia2020detection} used Faster RCNN to classify astronomical objects, including stars and stripe-like space targets. \cite{li2022bsc} and \cite{liu2023multi} adopted hierarchical U-Net-based networks to address stray light interference, providing valuable insights for the field. \cite{tao2023sdebrisnet} proposed SDebrisNet, a stripe detection network using CNN and attention mechanisms, though its efficacy under intense stray light remains unreported. 
In addition, advanced networks for complex backgrounds, such as ISNet\cite{zhang2022isnet}, ACM\cite{dai2021asymmetric}, AGPCNet \cite{zhang2021agpcnet}, DNANet\cite{li2022dense}, RDIAN\cite{sun2023receptive}, also rely on pixel-level labels.

Geometric and pattern-based approaches have laid the groundwork for SSTD, while data-driven methods promise improved detection. However, the field still grapple with the challenges of dataset availability, labeling costs, and model adaptability to the various space imaging conditions. Exploring weakly supervised learning and the potential of foundation models offer promising solutions to these challenges.

\section{AstroStripeSet Dataset}
\label{sec:dataset}

\begin{table}[h!]
    \centering
    \small 
    \setlength{\tabcolsep}{2pt} 
    \renewcommand{\arraystretch}{1.4}
    \begin{tabular}{|l|c|c|c|c|c|c|}
        \hline
        \textbf{Dataset} &  \textbf{Size} & \textbf{Diversity} & \textbf{Target} & \textbf{Label}  & \textbf{Public} \\ 
        \hline
        \cite{lin2021new} & 83  & One & Stripe &  $\times$  & $\times$   \\
        \cite{jiang2022automatic} & 200  & One & Stripe &  $\times$  & $\times$   \\ 
        \cite{liu2020topological} & 111  & One & Point &  $\times$  & $\times$  \\ 
        \cite{yao2022adaptive} & 100  & Two & Point &  $\times$  & $\times$   \\ 
        \cite{diprima2018efficient} & 400  & One & Point &  $\times$  & $\times$  \\ 
        \cellcolor{gray!25}AstroStripeSet (Ours) &  \cellcolor{gray!25} 1,500 &  \cellcolor{gray!25} Four & \cellcolor{gray!25} Stripe &   \cellcolor{gray!25} Three &  \cellcolor{gray!25} \checkmark\\
        \hline
    \end{tabular}

\caption{Comparison with existing space target datasets.}
\label{table:datasets}
\end{table}
The scarcity of open-source datasets has been a critical bottleneck in SSTD research, due to the high cost of obtaining numerous real space images for academic purposes. Inspired by solutions in other data-scarce domains like infrared small target detection \cite{li2022dense} and ship detection \cite{shermeyer2021rareplanes}, we developed AstroStripeSet, a large-scale synthetic dataset and the first comprehensive public dataset for this research community. Table~\ref{table:datasets} shows that AstroStripeSet surpasses previous in-house datasets in size, background diversity, and label type, facilitating a more robust and thorough approach to model training and testing. 

\begin{figure}[ht]
    \centering
    \begin{subfigure}{0.15\textwidth}  
        \includegraphics[width=\textwidth]{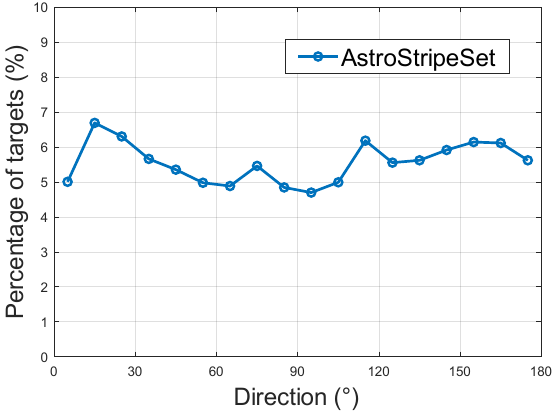}
        \caption{Target direction}
        \label{fig:target_direction}
    \end{subfigure}
    \hspace{1pt}  
    \begin{subfigure}{0.15\textwidth}
        \includegraphics[width=\textwidth]{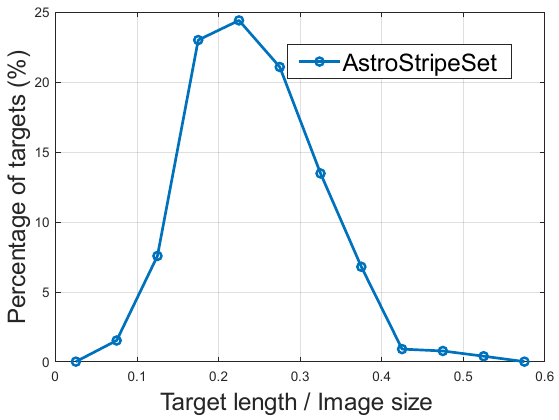}
        \caption{Target length}
        \label{fig:target_length}
    \end{subfigure}
    \hspace{1pt}  
    \begin{subfigure}{0.15\textwidth}
        \includegraphics[width=\textwidth]{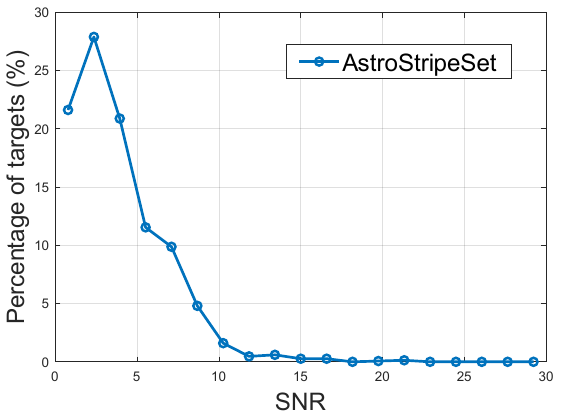}
        \caption{Target SNR}
        \label{fig:target_snr}
    \end{subfigure}
    \caption{Statistics for the AstroStripeSet dataset.}
    \label{fig:astrostripeset_statistics}
\end{figure}

AstroStripeSet includes a diverse array of space stray lights and sensor disturbances, such as reflections from the earth, moon, and sun, cosmic rays, and star background interference. This diversity is crucial for developing models that generalize across a wide range of real-world scenarios. Furthermore, the dataset features stripe-like targets with diverse positions, scales, orientations, and brightness distributions, providing a rich resource for developing and testing SSTD methods. Figure \ref{fig:astrostripeset_statistics} illustrates the distribution of the AstroStripeSet by stripe-like targets direction, length, and SNR. It offers three types of labels: single-point, precise pixel-level, and bounding box, supporting various research needs from basic detection to complex segmentation and localization tasks. The dataset consists of 1,500 pairs of raw and labeled images. It is divided into training, validation, and test sets, with fixed 1,000, 100, and 400 images, respectively. For detailed synthesis steps and image samples of our AstroStripeSet, please refer to the supplementary material. 

\section{Teacher-Student Label Evolution Framework using Single-Point Weak Supervision}
\label{sec:method}
\begin{figure*}[]
  \centering 
  \includegraphics[width=0.8\textwidth]{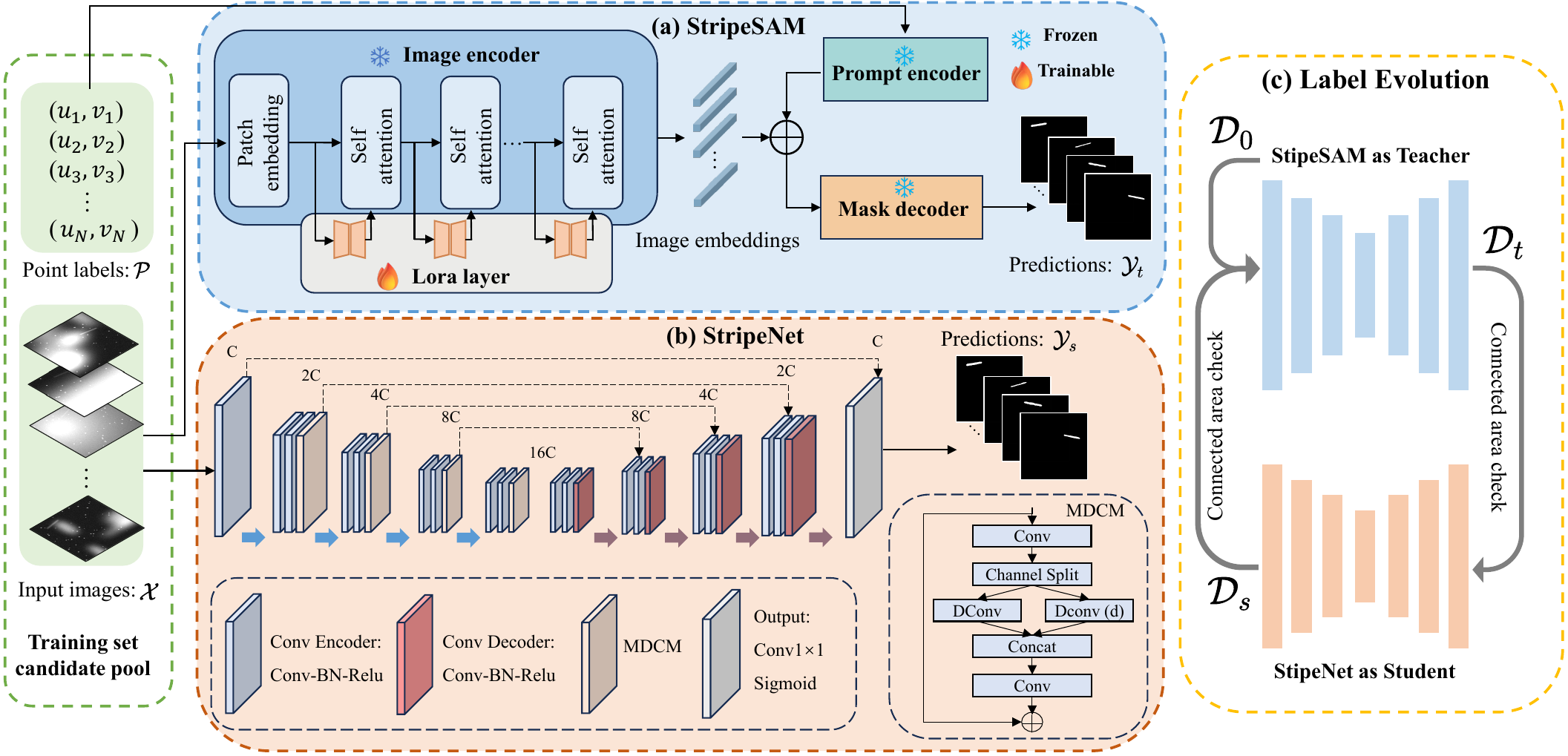}
  \caption{An overall framework for our teacher-student label evolution with single-point weak supervision.} 
  \label{fig:fig2} 
\end{figure*}

The teacher-student framework for SSTD task is depicted in Figure~\ref{fig:fig2}. The initial pseudo-labels are generated by the pre-trained SAM using the single-point labels. These pseudo-labels are then iterated between the teacher and student networks, initiating the label evolution process to achieve more accurate SSTD. This framework integrates two core components: the StripeSAM, a low-rank adaptation fine-tuned that functions as the teacher network, and the novel StripeNet, which serves as the student network. Furthermore,
Inspired by the linear distribution of stripe-like targets, we developed a customized loss function, GeoDice for StripeNet, to improve the learning of variable stripe-like target patterns.  

\subsection{Initial Pseudo-Labels from Pre-Trained SAM}
The initial step in our teacher-student framework for SSTD involves generating pseudo-labels using single-point target labels and the pre-trained SAM. This process begins with feeding the space images into the image encoder for generating image embeddings. Concurrently, single-point labels serve as point prompts for the stripe-like target areas, and are fed into the prompt encoder for prompt embeddings. The mask decoder generate initial pseudo-labels by mapping image and prompt embeddings to output tokens. 

To be specific, the input for the pre-trained SAM includes images $\mathcal{X} = \{ \mathbf{x}_i| i\in \{1,2, ...,N\}, \mathbf{x}_i \subseteq \mathbb{R}^{d} \}$, and point prompts $\mathcal{P} = \{(u_i, v_i)| i\in \{1,2, ...,N\}\}$, where $(u_i, v_i)$ is the single-point label for each image and $N$ is the total samples in the training set candidate pool.
Since the pre-trained SAM is designed for general segmentation tasks rather than specific space scenarios, its outputs tend to be very noisy, especially with faintly textured stripe-like targets. To overcome this limitation, we filter pseudo-labels using `connected area check'. Since we only use single-point annotated target regions as prompts for SAM, it should ideally produce only a single object segmentation result for the target connected region in each image. Given this, we count the number of objects by detecting connected regions, and any pseudo-labels with multiple objects are considered false positives and filtered out.
After that, the remaining segmentation masks are kept as the initial pseudo-labels, $\mathcal{Y}_0 = \{ \mathbf{y}_i| i\in \{1,2, ..., N_0\}, \mathbf{y}_i \subseteq \mathbb{R}^{d} \}$, $\mathbf{y}_i$ is the segment map and has the same size as the input image $\mathbf{x}_i$, the number of the selected images $N_0$ is usually much less than $N$. The inference is shown as Equation~\ref{equ:init}, in which the $\mathcal{F}$($\cdot$) is `connected area check' and $\mathcal{H}$($\cdot$) is the pre-trained SAM.
\begin{equation}
\label{equ:init}
\begin{aligned}
    \mathcal{Y}_0 = & \left\{ \mathcal{F} \left( \mathcal{H} \left( \mathbf{x}_i, u_i, v_i \right) \right) \mid \right. \\
    & \left. \mathbf{x}_i \in \mathcal{X}, \, (u_i, v_i) \in \mathcal{P}, \, i \in \{1, 2, \ldots, N\} \right\}
\end{aligned}
\end{equation}

With the filtered input image $\mathcal{X}_0$, point prompt $\mathcal{P}_0$, and pseudo-labels $\mathcal{Y}_0$, we can construct our initial training dataset, $\mathcal{D}_0 = (\mathcal{X}_0, \mathcal{P}_0, \mathcal{Y}_0) = \{(\mathbf{x}_i, (u_i, v_i), \mathbf{y}_i)\}^{N_0}_{1}$. This dataset is then fed into the subsequent teacher model to initiate the loop of label evolution.

\subsection{StripeSAM as the Fine-Tuning Teacher}

Despite its proficiency in zero-shot segmentation within high SNR scenarios, the pre-trained SAM struggles with stripe-like targets in intense stray light noise. This issue arises from its lack of training on noisy samples and prior knowledge of stripe-like patterns in such conditions. To address this issue, we fine-tune the SAM model with the initially selected dataset $\mathcal{D}_0$ using low-rank adaptation (LoRA) \cite{hu2021lora}. This method incorporates a novel low-rank weight matrix within the image encoder, as shown in the Figure~\ref{fig:fig2}(a).
By updating only a minimal subset of parameters, the SAM can be fine-tuned efficiently. 
The fine-tuned SAM, referred to hereafter as StripeSAM, serves as the teacher network in the label evolution framework. The original weight parameters of StripeSAM are frozen, while the low-rank matrices are trained using the selected dataset $\mathcal{D}_0$. By doing this, we find that StripeSAM not only demonstrates superior label quality over the pre-trained SAM but also enriches the training dataset with additional pairs of faintly textured labeled images.

Specifically, we use the StripeSAM $\mathcal{G}$(.) to infer all images from the training set candidate pool to generate the new pseudo-labels, as shown in Equation~\ref{equ:teacher}. These labels go through the `connected area check' again to filter out noisy labels, and the remained input image $\mathcal{X}_t$, pseudo-labels $\mathcal{Y}_t$, and their mass center points of target area $\mathcal{P}_t$ are used to train the following student network. Up to here, we obtain our new training dataset, $\mathcal{D}_t = (\mathcal{X}_t, \mathcal{P}_t, \mathcal{Y}_t) = \{(\mathbf{x}_i, (u_i, v_i), \mathbf{y}_i)\}^{N_t}_{1}$, where $N_t$ is the number of samples selected by StripeSAM. 
\begin{equation}
\label{equ:teacher}
\begin{aligned}
    \mathcal{Y}_t = & \left\{ \mathcal{F} \left( \mathcal{G} \left( \mathbf{x}_i, u_i, v_i \right) \right) \mid \right. \\
    & \left. \mathbf{x}_i \in \mathcal{X}_0, \, (u_i, v_i) \in \mathcal{P}_0, \, i \in \{1, 2, \ldots, N_0\} \right\}
\end{aligned}
\end{equation}

\subsection{StripeNet as the Evolving Student}

To enhance the stripe-like pattern recognition of the StripeSAM during label evolution, we introduce StripeNet, a student network with a lightweight encoder-decoder architecture, as shown in Figure~\ref{fig:fig2}(b). The encoder phase consists of many multi-scale dilated convolution modules (MDCM) and is aimed at heightening the network's sensitivity to stripe-like targets across varying scales. The introduction of StripeNet not only refines the quality of pseudo-labels but also enlarges the repository of data pairs with faintly textured labels. The dataset  $\mathcal{D}_t$ from the teacher network is used to train this student network. After that, the model is used to generate the pseudo-labels $\mathcal{Y}_s$ for the next iteration, with the input image and point prompts denoted as $\mathcal{X}_s$, $\mathcal{P}_s$, respectively. The label generation is as Equation~\ref{equ:student}, in which $\mathcal{E}(\cdot)$ represents the StripeNet. Therefore, we can construct a new dataset, $\mathcal{D}_s = (\mathcal{X}_s, \mathcal{P}_s, \mathcal{Y}_s) = \{(\mathbf{x}_i, (u_i, v_i), \mathbf{y}_i)\}^{N_s}_{1}$, where $N_s$ is the number of samples selected by StripeNet. This enriched dataset is then cycled back to StripeSAM.
\begin{equation}
\label{equ:student}
\begin{aligned}
    \mathcal{Y}_s = & \left\{ \mathcal{F} \left( \mathcal{E} \left( \mathbf{x}_i, u_i, v_i \right) \right) \mid \right. \\
    & \left. \mathbf{x}_i \in \mathcal{X}_t, \, (u_i, v_i) \in \mathcal{P}_t, \, i \in \{1, 2, \ldots, N_t\} \right\}
\end{aligned}
\end{equation}

To conclude, as shown in Figure \ref{fig:fig2}(c), our label change process is initialized from pre-trained SAM using single-point labels, and then iteratively loops in the teacher-student framework to enhance pseudo-label quality, which is called the label evolution. We find that more and more pseudo-labels will be exploited as the label evolution progresses. 

\subsection{ GeoDice Loss Function}
Stripe-like space targets typically resemble straight line shapes and are distributed along specific directions. Based on this observation, we introduce a new loss function called GeoDice. 
As shown in Equation~\ref{equ:geoloss}, it consists of geometric alignment loss $\mathcal{L}_g $ and weighted dice loss $\mathcal{L}_d$ \cite{li2021combined}. $\mathcal{L}_g$ is responsible for distinguishing fine-grained features between stripe-like targets and noise in local regions, while $\mathcal{L}_d$ focuses on controlling the overall segmentation accuracy. 

The GeoDice loss enhances object segmentation by integrating geometric direction information from straight line shapes, yielding segmentation results that more closely resemble true stripes. To implemente this loss, we first divide the binary prediction map into connected regions and then calculate the longest straight line in each connected region. To simplify straight line determination, we adopt the farthest point sampling algorithm to determine the two endpoints of the longest straight line from each connected area in the prediction map and the pseudo label. Then, the angle difference of the straight line between prediction and pseudo label is calculated and multiplied by the Sigmoid output of the predicted image to ensure its differentiability, and angle differences are averaged as $\mathcal{L}_g $. The specific equation for $\mathcal{L}_g $ is:

\begin{equation}
\begin{aligned}
    \mathcal{L}_g = & \frac{1}{N_b} \sum_{i=1}^{N_b} \eta ( \frac{1}{M_i} \sum_{j=1}^{M_i} \min(\angle(p_j, g_i),  \\
    & \left. \pi - \angle(p_j, g_i)) \times \sigma(p_j) \right)
\end{aligned}
\end{equation}
where $ N_b $ is the number of images in a batch, $ M_i $ is the number of connected regions in the $i$-th image, $\sigma(\cdot)$ is the sigmoid function, $\eta(\cdot)$ is the normalization function, and $\angle(p_j, g_i)$ is the angle between the $j$-th connected predicted area $p_j$ and pseudo label $g_i$. Finally, our GeoDice is expressed in Equatation~\ref{equ:geoloss}, with $\alpha$ and $\lambda$ as weighting values.

\begin{equation}
    \mathcal{L}_{gd} = \alpha \times \mathcal{L}_g + \lambda \times \mathcal{L}_d   
    \label{equ:geoloss}
\end{equation}

\section{Experiments}\label{sec:exper}

\setlength{\tabcolsep}{1mm}
\begin{table*}[h!]
    \centering
    \small
    \begin{tabularx}{\textwidth}{|c| c|Y Y Y c|Y Y Y c|Y Y Y c|Y Y Y c|} 
        \hline
        \multirow{2}{*}{\textbf{Type}} & \multirow{2}{*}{\textbf{Method}} & \multicolumn{4}{c|}{\textbf{Sun Light}} & \multicolumn{4}{c|}{\textbf{Moon Light}} & \multicolumn{4}{c|}{\textbf{Earth Light}} & \multicolumn{4}{c|}{\textbf{Mixed Light}} \\  \cline{3-18} 
        & & \textbf{Dice} & \textbf{mIoU} & 
        \textbf{P}\textsubscript{\textbf{d}} & \textbf{F}\textsubscript{\textbf{a}} & 
        \textbf{Dice} & \textbf{mIoU} & 
        \textbf{P}\textsubscript{\textbf{d}} & \textbf{F}\textsubscript{\textbf{a}} & 
        \textbf{Dice} & \textbf{mIoU} & 
        \textbf{P}\textsubscript{\textbf{d}} & \textbf{F}\textsubscript{\textbf{a}} & 
        \textbf{Dice} & \textbf{mIoU} & 
        \textbf{P}\textsubscript{\textbf{d}} & \textbf{F}\textsubscript{\textbf{a}} \\
        \hline
        \multirow{3}{*}{\textbf{Zero}} 
        & SAM-vit\_b  & 25.85& 22.05& 26.0& $>10^2$& 42.61& 37.81& 46.0& $>10^2$& 38.96& 34.55& 42.0& $>10^2$& 38.48& 33.37& 39.0& $>10^2$ \\ 
        & SAM-vit\_l & 35.49& 31.20& 37.0& $>10^2$& 51.26& 46.20& 54.0& $>10^2$& 44.76& 39.67& 46.0& $>10^2$& 39.23& 34.71& 40.0&$>10^2$ \\ 
        & SAM-vit\_h & 42.42& 37.02& 44.0& $>10^2$& 56.23& 50.37& 60.0& $>10^2$& 53.16& 46.96& 57.0& $>10^2$& 43.54& 38.44& 44.40&$>10^2$ \\ \hline
        \multirow{8}{*}{\textbf{Full}} & 
        UNet & 90.60& 83.75& 97.0& 0.97& 89.15& 82.76& 95.0& 0.95& 90.04& 84.01& 95.0& 0.92& 87.16& 80.21& 90.0& 1.39 \\ 
        & ACM & 81.70& 72.49& 90.0& 1.57& 80.99& 72.90& 88.0& 1.49& 81.59& 73.0& 89.0& 1.54& 80.97& 72.02& 87.0& 1.86 \\ 
        & DNANet & 89.85& 82.16& 98.0& 1.05& 88.93& 80.44& 94.0& 1.04& 89.51& 81.48& 96.0& 1.09& 84.48& 76.62& 88.0& 1.64 \\ 
        & RDIAN & 87.34& 78.89& 93.0& 1.30& 86.91& 78.82& 94.0& 1.20& 88.29& 80.45& 95.0& 1.17& 84.23& 75.74& 86.0& 1.73 \\ 
        & AGPCNet & 86.83& 78.43& 94.0& 1.32& 86.67& 78.44& 95.0& 1.24& 88.39& 80.33& 94.0& 1.21& 83.74& 75.35& 88.0& 1.70 \\ 
        & MResUNet & 26.51& 21.42& 0.23& 4.37& 40.46& 34.12& 40.0& 3.38& 38.31& 32.69& 38.0& 3.74& 37.73& 31.79& 35.0&4.50 \\ 
        & UCTransNet  & 91.11& 84.38& 98.0& 0.95& 91.46& 85.21& 99.0& 0.84& 91.52& 85.24& 98.0& 0.88& 90.58& 83.78& 98.0&1.14 \\        
        & StripeNet  & 92.81& 86.98& 99.0& 0.76& 90.87& 85.18& 97.0& 0.75& 90.97& 85.28& 97.0& 0.82& 89.91& 83.26& 96.0&1.06 \\ \hline          
        \multirow{4}{*}{\textbf{Weak}}
        & Se-StripeNet & 74.75& 64.83& 77.0& 2.06& 79.10& 70.13& 82.0& 1.67& 79.53& 70.52& 86.0& 1.79& 76.67& 67.35& 80.0& 2.28 \\ 
        & Se-StripeSAM & 77.49& 66.69& 81.0& 0.50& 79.69& 70.03& 84.0& 1.02& 81.22& 71.67& 89.0& 0.96& 75.72& 65.67& 77.0& 1.25 \\ 
        & \cellcolor{gray!25}Co-StripeNet & \cellcolor{gray!25}84.47 & \cellcolor{gray!25}74.52& \cellcolor{gray!25}92.0 & \cellcolor{gray!25}1.63& \cellcolor{gray!25}84.74 & \cellcolor{gray!25}75.23& \cellcolor{gray!25}91.0 & \cellcolor{gray!25}1.47& \cellcolor{gray!25}84.40 & \cellcolor{gray!25}75.31& \cellcolor{gray!25}92.0& \cellcolor{gray!25}1.56 & \cellcolor{gray!25}81.18 & \cellcolor{gray!25}71.72& \cellcolor{gray!25}84.0 & \cellcolor{gray!25}2.02 \\ 
        & \cellcolor{gray!25}Co-StripeSAM & \cellcolor{gray!25}85.50&\cellcolor{gray!25}75.44& \cellcolor{gray!25}96.0&\cellcolor{gray!25}0.02&\cellcolor{gray!25}85.93&\cellcolor{gray!25}76.35&\cellcolor{gray!25}97.0&\cellcolor{gray!25}0.02&\cellcolor{gray!25}86.95&\cellcolor{gray!25}77.81&\cellcolor{gray!25}97.0&\cellcolor{gray!25}0.03&\cellcolor{gray!25}85.28&\cellcolor{gray!25}75.75&\cellcolor{gray!25}93.0&\cellcolor{gray!25}0.16 \\ 
        \hline
    \end{tabularx}
\caption{Comparison with different SOTA networks, assessed using Dice $\uparrow$(\%), mIoU $\uparrow$(\%), $P_d$ $\uparrow$(\%), and $F_a$ $\downarrow$($×10^{-3}$) metrics.}
\label{table:table2}
\end{table*}

\begin{figure*}[h!]
  \centering 
  \includegraphics[width=\linewidth]{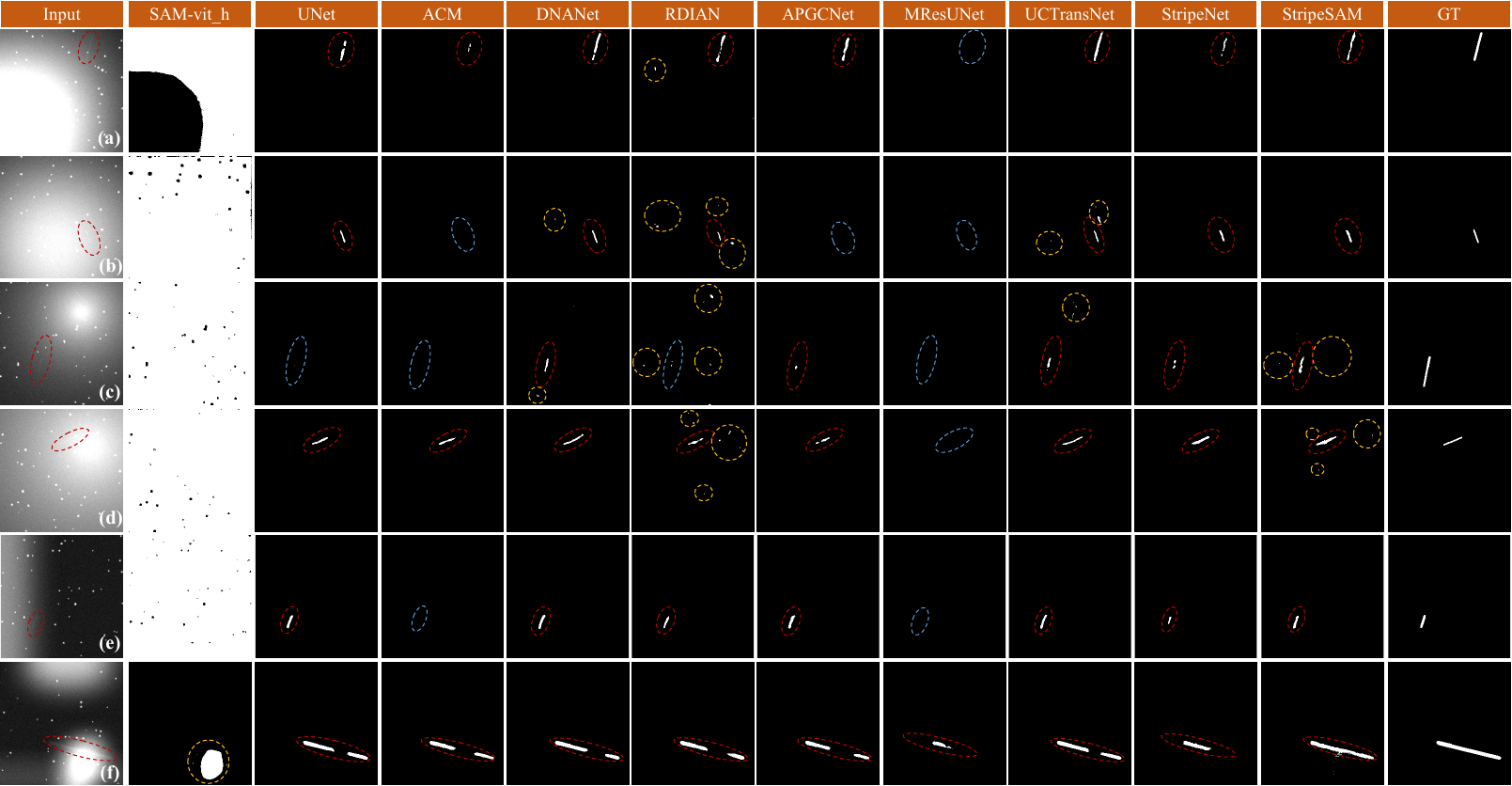}
  \caption{Visual comparison of our method with the SOTA fully supervised network on the upcoming AstroStripeSet. (Detected true targets, miss detections, and false alarms are highlighted in red, blue, and yellow, respectively).} 
  \label{fig:fig3} 
\end{figure*}

\subsection{Dataset and Evaluation Metrics}

\noindent\textbf{Dataset:}
Due to the unavailability of a large public dataset for SSTD with deep learning methods, we evaluated the algorithm using our up-coming AstroStripeSet. We also assessed the model's generalization ability using a diverse set of real-world space images. This includes real background images captured by our on-orbit cameras and ground-based telescopes, online sources, and images from the published papers \cite{lin2021new,li2022bsc}. These test data, featuring a variety of space imaging scenarios, challenge the method’s capacity to accurately detect stripe-like targets.


\noindent\textbf{Evaluation Metrics:}
To quantitatively assess our proposed framework, we use evaluation metrics including mean Intersection over Union (mIoU) \cite{dan2023dynamic} and Dice coefficient \cite{deng2018learning} for pixel-level accuracy, along with detection rate ($P_d$) and false alarm rate ($F_a$) for target-level performance. 
The mIoU and Dice metrics are particularly effective in assessing the algorithm's segmentation precision in identifying details within the targets area. Furthermore, $P_d$ quantifies the algorithm's success in correctly identifying stripe-like targets, and $F_a$ calculates the rate of incorrect detections, providing insight into the algorithm's sensitivity and specificity in distinguishing true targets from false alarms. In particular, targets are considered successfully detected with an IoU value greater than 50\%, while pixels outside the target region are classified as false alarms.

\subsection{Results and Discussion}

\noindent\textbf{Quantitative Results:}
Traditional methods based on geometric \cite{jiang2022space,jiang2022automatic,liu2020space1} and pattern \cite{sara2017faint,lin2021new} information struggle in scenarios with variable and intense stray light, making them unsuitable for direct comparison. Consequently, our evaluation primarily targets SOTA fully supervised deep learning models, such as UNet \cite{ronneberger2015u}, ACM \cite{dai2021asymmetric}, DNANet \cite{li2022dense}, AGPCNet \cite{zhang2021agpcnet}, RDIAN\cite{sun2023receptive}, MResUnet \cite{ibtehaz2020multiresunet}, and UCTransNet \cite{wang2022uctransnet}. We also compare our method with three pre-trained SAMs \cite{kirillov2023segment} employing single-point prompts, denoted SAM-vit\_b, SAM-vit\_l and SAM-vit\_h. Furthermore, we compare self-trained StripeSAM and StripeNet under single-point supervision, named Se-StripeSAM and Se-StripeNet, to show the significant benefits of our co-evolving teacher-student framework. Unless otherwise stated, StripeSAM and StripeNet are jointly trained using the teacher-student label evolution framework, referred to as Co-StripeSAM and Co-StripeNet.

Table~\ref{table:table2} presents a comprehensive quantitative comparison between the proposed framework and existing SOTA techniques. The analysis reveals that the performance of our teacher-student network significantly surpasses that of pre-trained single-point prompt SAM models across all four categories of challenging test images. This enhancement is not only in terms of segmentation accuracy but also addresses the challenge of pixel-level manual labeling, achieving a performance level on par with fully supervised network methods. 
In the fully supervised setting, StripeNet outperforms other networks significantly, offering faster inference and fewer parameters than the top-performing UCTransNet. This demonstrates that the customized StripeNet, as a student,  enriches StripeSAM with deeper insights into stripe patterns during the label evolution process, leading to more effective pseudo-labels generation for faintly textured stripe-like targets. This ongoing enhancement boosts the segmentation performance for low-SNR stripe-like targets. 

In the weakly supervised setting, Co-StripeSAM and Co-StripeNet significantly outperform Se-StripeSAM and Se-StripeNet across all metrics, and achieve performance comparable to fully supervised methods. Such advancements are attributed to the innovative implementation of a new teacher-student label evolution framework and the introduction of a customized GeoDice loss function, which collectively contribute to the superior performance of our approach. 

\noindent \textbf{Qualitative Assessment:}
This section examines the adeptness of our StripeSAM and StripeNet to show the detection of targets in low SNR scenarios. This is elucidated through a visual comparison in Figure~\ref{fig:fig3}, which compares them with the SOTA networks for detecting various stripe-like targets under various stray light conditions. The pre-trained SAM basically loses its ability to segment stripe-like targets under low SNR and requires manual prompts. While RDIAN can detect targets at low SNRs, it often produces many false positives. In contrast, StripeSAM performs impressively, matching the fully supervised UCTransNet, particularly in retaining the intricacies of stripe details throughout the spectrum of challenging images. Moreover, StripeNet's excellent false alarm suppression allows it automatically provides StripeSAM with a single-point target prompt during inference, eliminating the need for manual input and improving its practical usability.
The proficiency of StripeSAM is evident, as it consistently identifies true targets (marked in red), minimizes misses (marked in blue), and avoids false alarms (marked in yellow) with notable accuracy. Therefore, our findings substantiate the potential of StripeSAM as a superior alternative to the fully supervised SOTA methods in the context of stripe-like target detection amidst low SNR.

\subsection{Ablation Study}

\noindent\textbf{Comparative Analysis of Loss Functions:} 
Table~\ref{table:table4} shows the results of different loss for the first iteration of StripeNet. We compare standard segmentation loss functions and their combinations, such as weighted dice loss $\mathcal{L}_d$ \cite{li2021combined} and softIoU loss $\mathcal{L}_s$ \cite{li2022dense}, each paired with focal loss $\mathcal{L}_f$ \cite{lin2017focal} at a 1:1 ratio. We also include the SOTA shape-aware clDice loss \cite{shit2021cldice}, $\mathcal{L}_{cd}$. The results show that our customized shape-aware GeoDice loss ($\mathcal{L}_g+\mathcal{L}_d$) effectively uses prior information about stripe-like targets, significantly improving detection rates with minimal false positives while maintaining accuracy.

\setlength{\tabcolsep}{1mm}
\begin{table*}[t!]
    \centering
    \small
    \begin{tabularx}{\textwidth}{|c|Y Y Y Y|Y Y Y Y|Y Y Y Y|Y Y Y Y|}
        \hline
        \multirow{2}{*}{\textbf{Loss}} & \multicolumn{4}{c|}{\textbf{Sun Light}} & \multicolumn{4}{c|}{\textbf{Moon Light}} & \multicolumn{4}{c|}{\textbf{Earth Light}} & \multicolumn{4}{c|}{\textbf{Mixed Light}} \\ \cline{2-17}

        & \textbf{Dice} & \textbf{mIoU} & \textbf{P\textsubscript{d}} & \textbf{F\textsubscript{a}} & 
        \textbf{Dice} & \textbf{mIoU} & \textbf{P\textsubscript{d}} & \textbf{F\textsubscript{a}} & 
        \textbf{Dice} & \textbf{mIoU} & \textbf{P\textsubscript{d}} & \textbf{F\textsubscript{a}} & 
        \textbf{Dice} & \textbf{mIoU} & \textbf{P\textsubscript{d}} & \textbf{F\textsubscript{a}} \\ \hline
     
        \textbf{$\mathcal{L}_s$} & 68.73& 55.24& 68.0& 3.96& 70.45& 57.22& 68.0& 3.62& 72.39& 59.68& 75.0& 3.48& 68.53& 56.79& 71.0& 3.83\\ 
        \textbf{$\mathcal{L}_d$} & 75.98& 66.54& 78.0& 2.01& 76.60& 67.90& 78.0& 1.80& 77.65& 68.89& 83.0& 1.95& 72.89& 64.73& 73.0& 2.48\\ 
        \textbf{$\mathcal{L}_d+\mathcal{L}_f$} & 76.85& 67.19& 81.0& 2.06& 78.56& 69.63& 84.0& 1.73& 80.16& 70.60& 85.0& 1.89& 73.98& 64.99& 79.0& 2.40\\ 
        \textbf{$\mathcal{L}_s+\mathcal{L}_f$} & 72.99& 64.67& 79.0& 2.14& 76.19& 68.09& 83.0& 1.83& 78.91& 70.36& 84.0& 1.90& 73.10& 65.06& 79.0& 2.39\\ 
        \textbf{$\mathcal{L}_{cd}$} & 78.22& 69.15& 82.0& 1.98& 78.36& 70.27& 82.0& 1.64 & 81.07& 72.63& 88.0& 1.92& 73.16& 65.84& 77.0& 2.73 \\ 
        
        \cellcolor{gray!25}\textbf{$\mathcal{L}_g+\mathcal{L}_d$}    & \cellcolor{gray!25}82.05 & \cellcolor{gray!25}71.76& \cellcolor{gray!25}89.0& \cellcolor{gray!25}1.78& \cellcolor{gray!25}81.42 & \cellcolor{gray!25}72.04& \cellcolor{gray!25}88.0 & \cellcolor{gray!25}1.60& \cellcolor{gray!25}80.56 & \cellcolor{gray!25}71.24& \cellcolor{gray!25}87.0 & \cellcolor{gray!25}1.85& \cellcolor{gray!25}73.44 & \cellcolor{gray!25}64.64& \cellcolor{gray!25}78.0 & \cellcolor{gray!25}2.40 \\ \hline
    \end{tabularx}%
\caption{Comparison results of different loss functions in StripeNet.}
\label{table:table4}
\end{table*}

\setlength{\tabcolsep}{1mm}
\begin{table*}[h!]
    \centering
    \small
    \begin{tabularx}{\textwidth}{|c|c|Y Y Y Y|Y Y Y Y|Y Y Y Y|Y Y Y Y|}
    \hline
    \multirow{2}{*}{\textbf{Iteration}} & \multirow{2}{*}{\textbf{Method}} & \multicolumn{4}{c|}{\textbf{Sun Light}} & \multicolumn{4}{c|}{\textbf{Moon Light}} & \multicolumn{4}{c|}{\textbf{Earth Light}} & \multicolumn{4}{c|}{\textbf{Mixed Light}} \\ \cline{3-18} 
       & & \textbf{Dice} & \textbf{mIoU} & \textbf{P\textsubscript{d}} & \textbf{F\textsubscript{a}} & \textbf{Dice} & \textbf{mIoU} & \textbf{P\textsubscript{d}} & \textbf{F\textsubscript{a}} & \textbf{Dice} & \textbf{mIoU} & \textbf{P\textsubscript{d}} & \textbf{F\textsubscript{a}} & \textbf{Dice} & \textbf{mIoU} & \textbf{P\textsubscript{d}} & \textbf{F\textsubscript{a}} \\ \hline
        \multirow{2}{*}{\textbf{1}} 
        & StripeNet & 75.08& 65.40& 79.0& 2.07& 75.62& 67.44& 78.0& 1.76& 77.47& 68.44& 83.0& 1.93& 71.45& 62.69& 71.0& 2.50\\
        & \cellcolor{gray!25}StripeSAM & \cellcolor{gray!25}61.67& \cellcolor{gray!25}53.35& \cellcolor{gray!25}63.0& \cellcolor{gray!25}4.10& \cellcolor{gray!25}70.52& \cellcolor{gray!25}62.76& \cellcolor{gray!25}75.0& \cellcolor{gray!25}0.25& \cellcolor{gray!25}67.91& \cellcolor{gray!25}60.55& \cellcolor{gray!25}74.0& \cellcolor{gray!25}2.75& \cellcolor{gray!25}61.49& \cellcolor{gray!25}53.62& \cellcolor{gray!25}61.0& \cellcolor{gray!25}5.31\\ 
        \hline
        \multirow{2}{*}{\textbf{2}} 
        & StripeNet  & 77.71& 68.41& 81.0& 1.88& 81.02& 71.68& 86.0& 1.62& 82.51& 73.43& 89.0& 1.59& 77.36& 68.58& 82.0& 2.16\\ 
        & \cellcolor{gray!25}StripeSAM  & \cellcolor{gray!25}82.04& \cellcolor{gray!25}71.58& \cellcolor{gray!25}88.0& \cellcolor{gray!25}0.04& \cellcolor{gray!25}84.27& \cellcolor{gray!25}74.76& \cellcolor{gray!25}91.0& \cellcolor{gray!25}0.06& \cellcolor{gray!25}84.78& \cellcolor{gray!25}75.43& \cellcolor{gray!25}89.0& \cellcolor{gray!25}0.08& \cellcolor{gray!25}81.85& \cellcolor{gray!25}71.96& \cellcolor{gray!25}85.0& \cellcolor{gray!25}0.10\\ 
        \hline
        \multirow{2}{*}{\textbf{3}} 
        & StripeNet  & 84.47& 75.52& 92.0& 1.63& 84.74& 75.23& 91.0& 1.47& 84.40& 75.31& 92.0& 1.56& 81.18& 71.72& 84.0& 2.02\\ 
        & \cellcolor{gray!25}StripeSAM & \cellcolor{gray!25}85.45& \cellcolor{gray!25}75.44& \cellcolor{gray!25}96.0& \cellcolor{gray!25}0.02& \cellcolor{gray!25}85.93& \cellcolor{gray!25}76.35& \cellcolor{gray!25}97.0& \cellcolor{gray!25}0.02& \cellcolor{gray!25}86.94& \cellcolor{gray!25}77.81& \cellcolor{gray!25}97.0& \cellcolor{gray!25}0.03& \cellcolor{gray!25}85.28& \cellcolor{gray!25}75.75& \cellcolor{gray!25}93.0& \cellcolor{gray!25}0.16\\ 
        \hline
    \end{tabularx}%
\caption{The impact of different iterations of teacher-student on detection results.}
\label{table:table5}
\end{table*}

\begin{figure}[h!]
  \centering 
  \includegraphics[width=\linewidth]{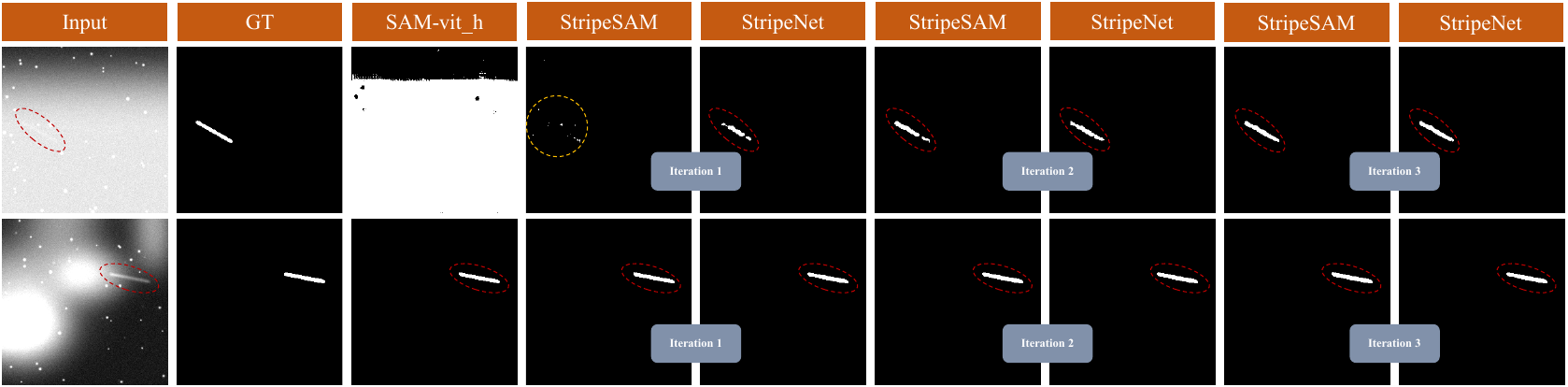}
  \caption{Results from three label evolution iterations.} 
  \label{fig:fig4} 
\end{figure}

\noindent\textbf{Effectiveness of Label Evolution:}
As noted previously, the pre-trained SAM can produce excellent results in high SNR scenarios, which also forms the basis for the success of our point-supervised framework. However, it struggles to segment faintly textured stripe-like targets in low SNR scenarios. Figure \ref{fig:fig4} shows that our label evolution approach improves the segmentation of stripe-like targets in low SNR scenarios. Table \ref{table:table5} presents the performance of StripeSAM and StripeNet for various iteration numbers, further highlighting the effectiveness of our label evolution strategy.

\begin{figure}[ht]
  \centering 
  \includegraphics[width=\linewidth]{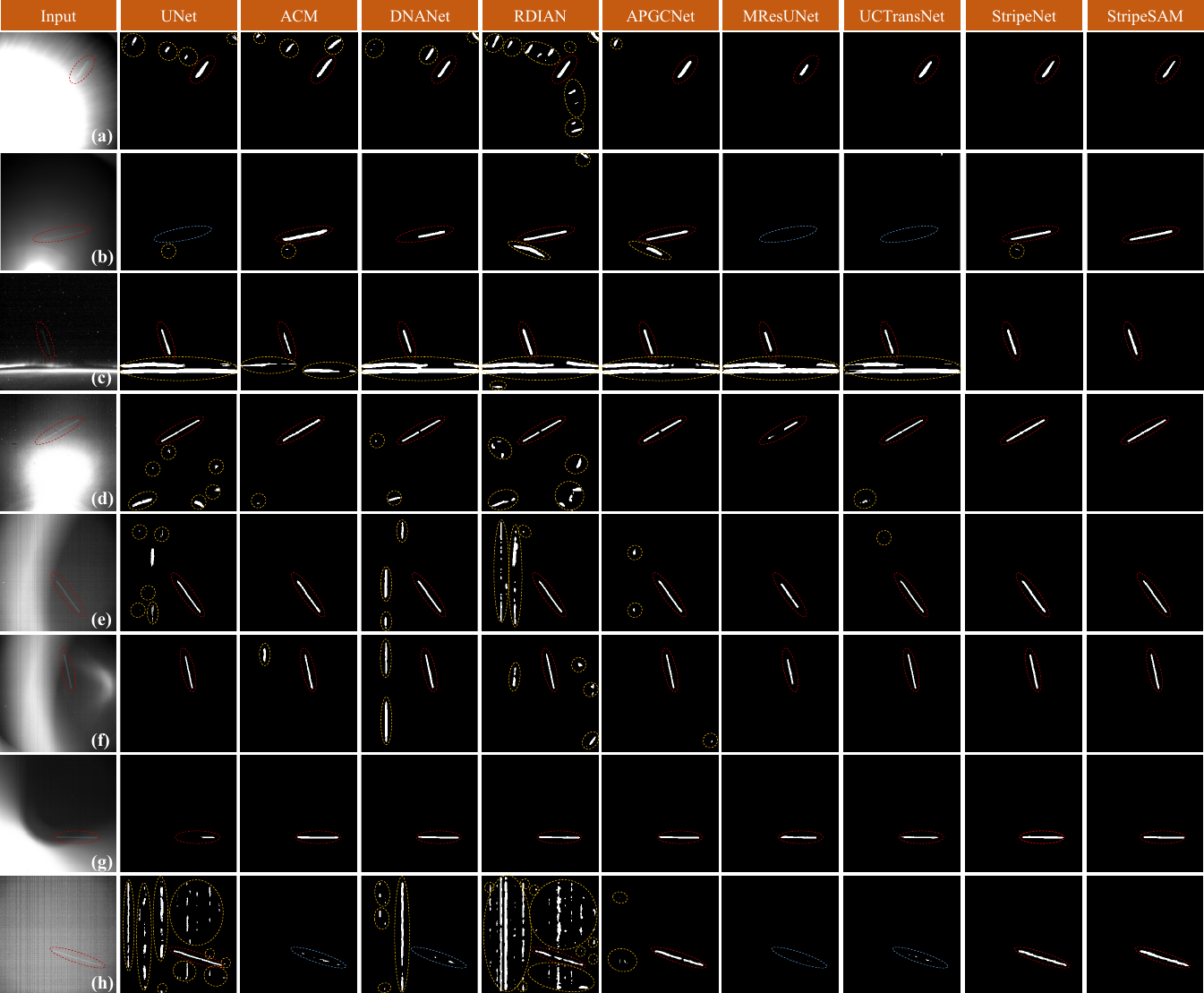}
  \caption{Visual comparison of detection under on-orbit and ground-based real-world images. } 
  \label{fig:fig5} 
\end{figure}

\noindent\textbf{Zero-Shot Generalization:} 
This section explores our model’s generalizability, evaluating its performance beyond our dataset and comparing it with other methods. Figure \ref{fig:fig5} shows the detection results of various approaches under real-world space stray light conditions, with Figures \ref{fig:fig5}(a)-\ref{fig:fig5}(d) taken by our on-orbit cameras, and Figures \ref{fig:fig5}(e)-\ref{fig:fig5}(h) taken by our ground-based telescopes. In particular, StripeSAM and StripeNet excel at accurately identifying stripe-like targets with minimal false positives. For more visual results on other datasets, please see the supplementary material.
Table \ref{table:table6} shows the average metrics of different methods on all real datasets. Our customized StripeNet and GeoDice loss allow StripeSAM to integrate more stripes knowledge, achieving zero-shot segmentation that outperforms fully supervised methods. This demonstration of effective zero-shot generalization highlights the strong adaptability of our framework. It also underscores the value of our AstroStripeSet dataset in preparing the network for real-world challenges.  

\begin{table}[h!]
\centering
\small
\begin{tabularx}{\columnwidth}{|c| c |Y Y Y Y|} 
    \hline
    \multirow{2}{*}{\textbf{Type}} & \multirow{2}{*}{\textbf{Method}} & \multicolumn{4}{c|}{\textbf{Average Metrics}}  \\  
    \cline{3-6} 
    & & \textbf{Dice}  &
    \textbf{mIoU}  & 
    \textbf{P\textsubscript{d}} & \textbf{F\textsubscript{a}}  \\
    \hline
    \multirow{8}{*}{\textbf{Full}} 
    & UNet & 42.05 & 30.78 & 23.53 & 11.45\\
    & ACM  & 62.59 & 50.85 & 52.94 & 4.45  \\
    & DNANet & 57.20 & 28.48 & 52.94 & 9.46 \\
    & RDIAN & 49.27 & 37.02 & 23.53 & 18.61  \\
    & APGCNet & 74.79 & 63.36 & 76.47 & 6.21\\
    & MResUNet  & 33.86 & 26.09 & 29.41 & 9.38  \\
    & UCTransNet & 56.05 & 46.90 & 58.82 & 6.49\\
    \hline
    \multirow{2}{*}{\textbf{Weak}} 
    & StripeNet & 82.56 & 71.81 & 82.36 & 2.15 \\
    & \cellcolor{gray!25}StripeSAM & \cellcolor{gray!25}89.66 & \cellcolor{gray!25}82.33 & \cellcolor{gray!25}100.0 & \cellcolor{gray!25}0.23 \\
    \hline
\end{tabularx}
\caption{Zero-shot generalization on real-world images.}
\label{table:table6}
\end{table}

\section{Conclusion}
\label{sec:conc}
In this study, we introduce AstroStripeSet, the inaugural dataset destined for open-source release in the stripe-like space target detection (SSTD) domain, establishing a new benchmark for future evaluations. We also present an innovative label evolution teacher-student model, employing single-point supervision alongside a custom GeoDice loss function, specifically designed for the identification of stripe-like targets under complex stray light conditions. Our comprehensive experiments demonstrate the method's exceptional performance in SSTD, offering a novel perspective within this field and expanding the scope of research in weakly supervised methods. Looking ahead, our goal is to explore the adaptation of our methodology to other segmentation challenges, incorporate other semi-supervised and unsupervised learning techniques, and investigate dynamic loss function adaptation and cross-domain transfer learning to enhance model versatility and applicability. Collecting more challenging real-space background images will also be a priority, further propelling SSTD research.

\bigskip

\bibliography{aaai25}


\end{document}